\documentclass[sigconf, screen]{acmart}

\usepackage{adjustbox}
\usepackage{multirow}
\AtBeginDocument{%
  }
\acmConference[Preprint]{Preprint}{Apr 08, 2026}{Arxiv}
\setcopyright{none}
\renewcommand\footnotetextcopyrightpermission[1]{}
\usepackage{adjustbox}
\usepackage{makecell}
\usepackage[table]{xcolor}
\begin{document}

\title{SurFITR: A Dataset for Surveillance Image Forgery \\Detection and Localisation}

\author{Qizhou Wang}
\affiliation{%
  \institution{The University of Melbourne}
  \city{Parkville}
  \country{Australia}
}
\email{mike.wang@unimelb.edu.au}

\author{Guansong Pang}
\affiliation{%
  \institution{Singapore Management University}
  % \city{Sin}
  \country{Singapore}
}
\email{gspang@smu.edu.sg}

\author{Christopher Leckie}
\affiliation{%
  \institution{The University of Melbourne}
  \city{Parkville}
  \country{Australia}
}
\email{caleckie@unimelb.edu.au}

\renewcommand{\shortauthors}{Wang et al.}
\newcommand{\mike}[1]{\textcolor{blue}{[mike]: #1}}

\begin{abstract}
  We present the $\textbf{Sur}$veillance $\textbf{F}$orgery $\textbf{I}$mage $\textbf{T}$est $\textbf{R}$ange (SurFITR), a dataset for surveillance-style image forgery detection and localisation, in response to recent advances in open-access image generation models that raise concerns about falsifying visual evidence. 
  Existing forgery models, trained on datasets with full-image synthesis or large manipulated regions in object-centric images, struggle to generalise to surveillance scenarios. This is because tampering in surveillance imagery is typically localised and subtle, occurring in scenes with varied viewpoints, small or occluded subjects, and lower visual quality.
  To address this gap, SurFITR provides a large collection of forensically valuable imagery generated via a multimodal LLM-powered pipeline, enabling semantically aware, fine-grained editing across diverse surveillance scenes. It contains over 137k tampered images with varying resolutions and edit types, generated using multiple image editing models. Extensive experiments show that existing detectors degrade significantly on SurFITR, while training on SurFITR yields substantial improvements in both in-domain and cross-domain performance. SurFITR is publicly available on 
  GitHub.\footnote{\url{https://github.com/mike-qz-wang/SurFITR}}.
\end{abstract}

\begin{teaserfigure}
\centering
  \includegraphics[width=1.0\textwidth]{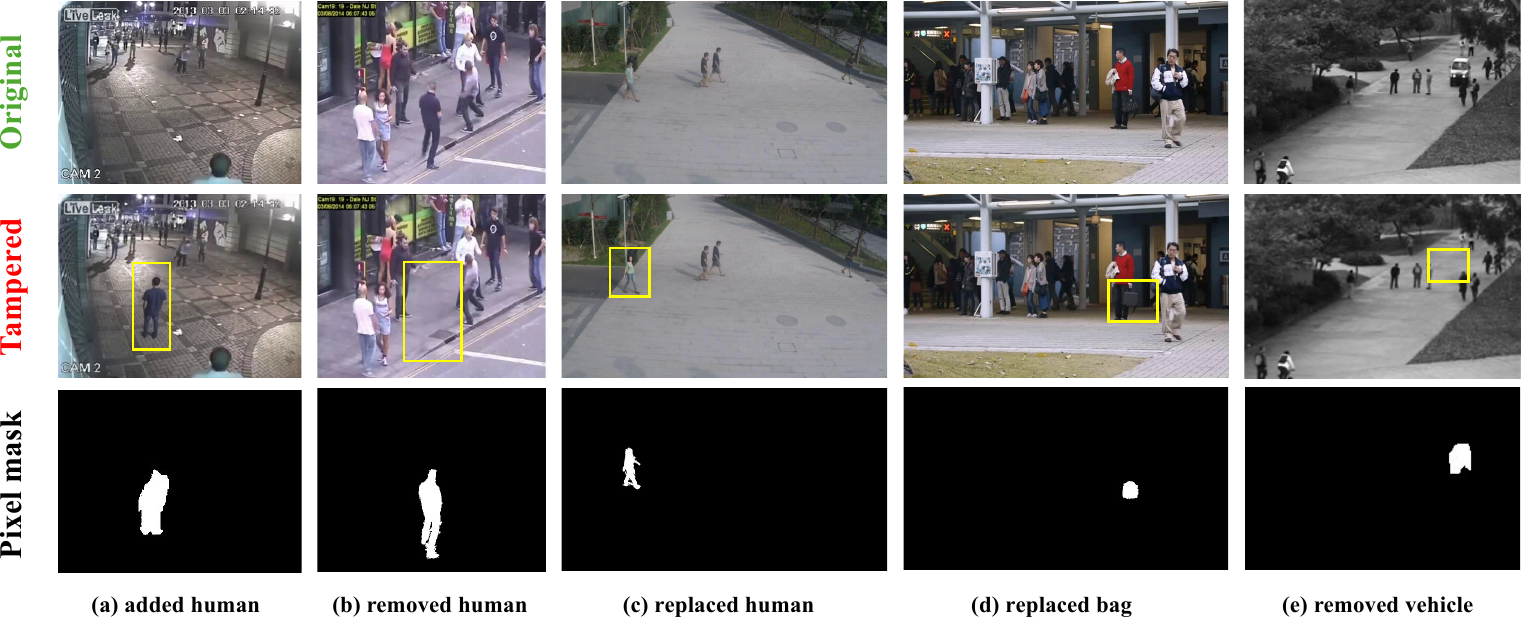}
  \caption{Visualisations from SurFITR showing realistic, fine grained tampering across diverse surveillance scenes. Top row: original images, middle row: tampered images (yellow boxes indicate manipulated regions), bottom row: edited pixel masks.}
  \label{fig:teaser}
\end{teaserfigure}


\maketitle

\section{Introduction}
Recent publicly available image generation models \cite{dall2, sd3.5, flux2024, wu2025Qwenimage, saharia2022imagen} can now achieve photorealistic quality comparable to some proprietary systems, enabling fine-grained and controllable edits. While these advances democratise creative tools, they also raise growing concerns about generative authenticity, as such models can be misused to falsify visual evidence or fabricate convincing misinformation. One example of a concerning threat this creates is the potential disruption of online reporting systems \cite{fbi_tips, crimestoppers}, where photos can be directly submitted.

\begin{figure}[t]
    \centering
    \includegraphics[width=0.7\linewidth]{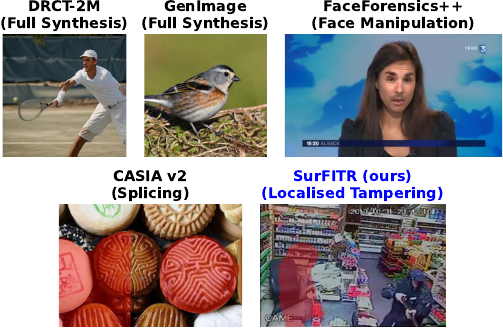}
    \vspace{-0.3cm}
    \caption{Comparison between SurFITR and prior datasets illustrating their differences. Red indicates manipulated regions, while absence denotes full generation or missing masks. Here, the SurFITR image removes the cashier, though the register was attended in the original image.}
    \label{fig:dset_comp}
    \vspace{-0.5cm}
\end{figure}

Despite extensive research of image forgery detection and localisation \cite{casia, psccnet, hifinet, trufor, drct, fakeshield}, these tasks remain a unique and underexplored challenge in surveillance-style imagery. 
Unlike the object-centric images used in existing benchmarks, surveillance imagery involves wider viewpoints, smaller or occluded subjects, and lower visual quality, with tampering that is often subtle and localised.
These differences weaken forensic cues and cause models trained on existing datasets to degrade under subtle manipulations in surveillance scenes. We illustrate these differences in Fig.~\ref{fig:dset_comp}. As a result, the lack of suitable datasets hinders both evaluation and the development of image forensic models for surveillance imagery.

To address this gap, we introduce the $\textbf{Sur}$veillance $\textbf{F}$orgery $\textbf{I}$mage $\textbf{T}$est $\textbf{R}$ange (SurFITR), a specialised dataset that enables forgery detection and localisation in real-world surveillance scenarios. It contains over 137k tampered images from six surveillance-style corpora consisting of diverse and forensically valuable scenes, covering four editing operations with both human and object manipulations, generated using five state-of-the-art (SOTA) image editing models. The images span a wide range of resolutions, include both colour and grayscale formats, and capture diverse activities. Beyond its focus on surveillance imagery and forensic relevance, SurFITR is distinguished by three key aspects: (1) an automated pipeline leveraging multimodal LLMs (MLLMs), open-world grounding models, and image generation models for semantically grounded, scene-specific editing, enabling the simulation of real-world forgery at scale, (2) semantically grounded, fine-grained tampering that reflects realistic editing across diverse scales, accompanied by precise pixel-level masks for training and evaluating localisation models, and (3) a consistent dataset structure that enables the study of cross-domain settings, including cross-scene, cross-model, and combined scenarios. We compare SurFITR with representative forgery datasets in Tab.~\ref{tab:dataset_comparison} across several dimensions: domain (Dom.), availability of pixel-level localisation masks (Loc.), inclusion of multiple editing types (Multi-Edit), inclusion of multiple generation models (Multi-Gen), context-aligned editing for each scene (Sem.), and diversity of source corpora (Div. Src.).

SurFITR serves as a valuable resource for both evaluation and training. Through extensive experiments, we show that current forensic methods trained on existing datasets, as well as pretrained MLLMs, struggle to detect and localise tampering in surveillance imagery, indicating a critical gap in existing benchmarks for training and evaluation.
When used for training, SurFITR-tuned models achieve significant gains in both in-domain and cross-domain settings, indicating forgery-discriminative, generalisable supervision from SurFITR.
Despite these gains, the tuned performance remains far from optimal. Through detailed analysis, we identify clear gaps in cross-domain detection and the localisation of subtle, fine-grained tampering. In particular, we observe that scene variation is a primary cause of instability in detection performance, while smaller manipulation regions significantly degrade localisation. These findings point to the need for specialised models that can learn scene-invariant cues and support fine-grained localisation. SurFITR provides a foundation for studying these challenges. Our contributions can be summarised as follows:
\begin{itemize}
    \item We introduce \textbf{SurFITR}, a dataset for surveillance-style image forgery detection and localisation, capturing realistic, fine-grained, and spatially localised tampering in complex real-world scenes.
    \item We develop a \textbf{MLLM-driven pipeline} for semantically grounded, scene-aware editing across diverse surveillance scenes, leveraging multiple SOTA image generation models that enable the study of cross-domain generalisation.
    \item We conduct extensive experiments showing that existing forensic models and MLLMs exhibit reduced performance in surveillance settings, while training on SurFITR yields substantial improvements, demonstrating its value as both a benchmark and a training resource.
\end{itemize}

\begin{table}[t]
\centering
\small
\begin{adjustbox}{width=\linewidth}
\begin{tabular}{l c c c c c c}
\toprule
\textbf{Dataset} & \textbf{Dom.} & \textbf{Loc.} & \textbf{Multi-Edit} & \textbf{Multi-Gen} & \textbf{Sem.} & \textbf{Div. Src.} \\
\midrule
DRCT-2M \cite{drct}          & Nat.  &  &  & \checkmark &  &  \\
GenImage \cite{genimage}        & Nat.  &  &  & \checkmark &  &  \\
FaceForensics++ \cite{faceforensics} & Face  &  & \checkmark &  &  &  \\
CASIA v2 \cite{casia}        & Nat.  &  & \checkmark &  &  &  \\
IMD2020 \cite{imd2020}        & Nat.  & \checkmark & \checkmark &  &  &  \\
\textbf{SurFITR} & Surv. & \checkmark & \checkmark & \checkmark & \checkmark & \checkmark \\
\bottomrule
\end{tabular}
\end{adjustbox}
\caption{SurFITR vs other representative datasets.}
\label{tab:dataset_comparison}
\vspace{-1cm}
\end{table}

\section{Related Work and Background}
\noindent \textbf{Image Forgery Datasets.} Existing datasets primarily focus on object-centric and face-centric scenarios. Early benchmarks such as CASIA v2 \cite{casia}, Columbia Splicing Dataset \cite{columbia}, and CoMoFoD \cite{comofod} focus on splicing and copy-move manipulations in object-centric images. Face-centric datasets such as FaceForensics++ \cite{faceforensics} and ForgeryNet \cite{he2021forgerynet} provide large-scale benchmarks but are limited to controlled facial scenarios. More recent datasets, including IDM2020 \cite{imd2020}, GenImage \cite{genimage}, DRCT-2M \cite{drct}, and DF2023 \cite{dolhansky2020deepfake}, leverage generative models to improve scale and diversity. However, these datasets are based on object-centric natural images and often involve hard, non-seamless edits that are not representative of real-world tampering, where manipulations are typically subtle and localised and occur in complex, heterogeneous scenes. \smallskip

\noindent \textbf{Forgery Detection and Localisation Methods.} Image forgery detection and localisation have been widely studied using CNNs and transformers. CNN-based methods \cite{psccnet, mvssnet, catnet, hifinet, drct} capture local artefacts, while transformers model global context for improved localisation \cite{objectformer, trufor, imlvit}. Large-scale detectors further explore data-driven classification \cite{drct}, and MLLMs enable zero-shot detection via visual–text reasoning \cite{gemini, qwen25vl, qwen3vl, fakeshield}. However, these methods are mainly developed on object-centric datasets and degrade under subtle, localised tampering in surveillance settings. \smallskip

\noindent \textbf{Image Generation Models.} Early generative models \cite{vae, gan, radford2015unsupervised, brock2018large, karras2019style} enabled high-fidelity synthesis but suffered from blur, instability, and mode collapse. Diffusion models, particularly DDPMs \cite{ho2020ddpm}, have emerged as a leading paradigm, with latent diffusion supporting scalable high-resolution generation \cite{rombach2022ldm}. Systems such as DALL-E 2 \cite{dall2}, Stable Diffusion, and Imagen \cite{saharia2022imagen}, followed by SDXL \cite{podell2024sdxl}, FLUX \cite{flux2024}, and Qwen-Image \cite{wu2025Qwenimage}, achieve highly realistic and controllable synthesis. However, they remain limited in instruction-driven localised editing due to weak spatial grounding.

\section{SurFITR Dataset}

\subsection{Dataset Overview}
As shown in Table~\ref{tab:struc}, SurFITR consists of two collections constructed under different generation settings: SurFITR-Base and SurFITR-Transfer. SurFITR-Base is generated using FLUX.1-Fill-dev \cite{flux2024} and serves as the primary training and evaluation set. SurFITR-Transfer is generated using four additional SOTA \cite{wu2025Qwenimage,sd3.5,zimage,flux-2-2025} image editing models (see Sec.~A.2.2 for model details) via LanPaint \cite{zheng2025lanpaint} and is designed to assess generalisation under cross-domain shifts, including both scene and generation variations (see Sec.~A.4 for usage details). It consists of two splits: Eval 1 and Eval 2, following the Base train and test scene splits, respectively, and are used for evaluation only

\begin{table}[t]
\centering
\begin{adjustbox}{width=0.8\linewidth}
\begin{tabular}{cccccc}
\toprule
Collections & Split & \# Real & \# Fake & Total & Verified Set \\
\midrule
\multirow{2}{*}{Base} & Train & 52,752  & 52,752 & 105,504 & -- \\
& Test  &  55,419  &  55,419  & 110,838 & 2,801  \\
\midrule
\multirow{2}{*}{Transfer} & Eval 1 & 14,963 & 14,963 & 29,926 & \multirow{2}{*}{1,501} \\
& Eval 2 & 15,022 & 15,022 & 30,044 & \\
\midrule
Total & & 137,439 & 137,439 & 276,312  & 5,065 \\
\bottomrule
\end{tabular}
\end{adjustbox}
\caption{SurFITR statistics.}
\label{tab:struc}
\vspace{-0.5cm}
\end{table}

\begin{figure}
    \centering
    \includegraphics[width=0.85\linewidth]{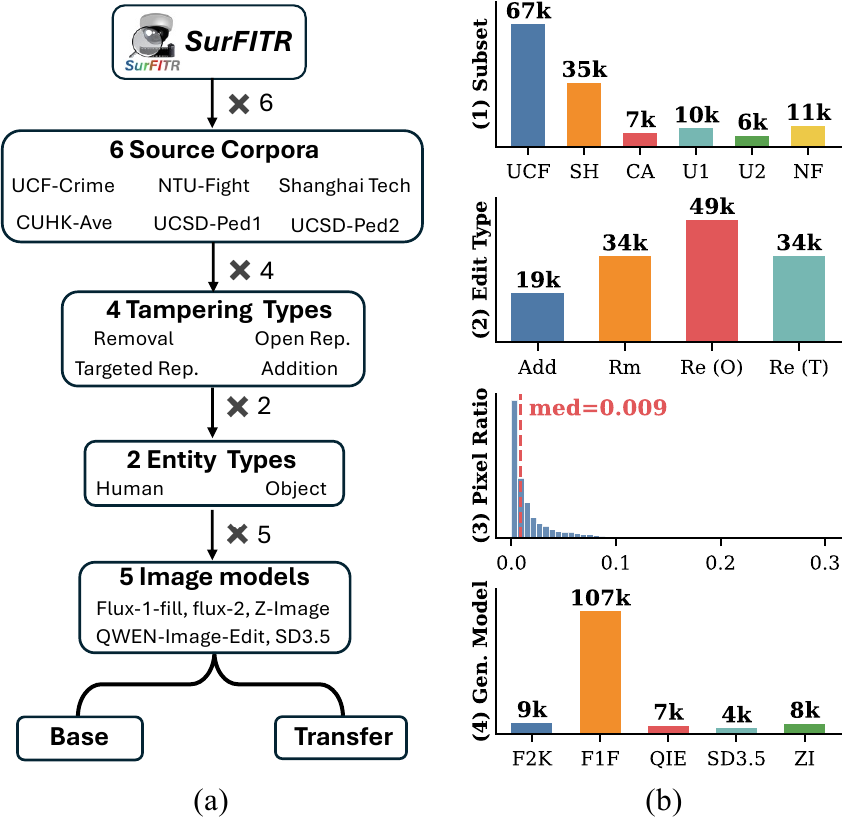}
    \vspace{-0.2cm}
    \caption{Overview of SurFITR. (a) Dataset structure illustrating the combinational design across source corpora, tampering types, entity types, and generation models. (b) Key statistics, including data distribution, edit types, manipulation scale, and generation models.}
    \label{fig:dset_stats}
    \vspace{-0.5cm}
\end{figure}

As shown in Fig. \ref{fig:dset_stats}, both collections are built upon six widely used surveillance datasets: UCF Crime~\cite{ucf}, NTU Fight~\cite{ntu_fight}, ShanghaiTech~\cite{shanghai}, CUHK Avenue \cite{cuhk}, UCSD Ped1 \cite{ucsd1}, and UCSD Ped2 \cite{ucsd2} (see Sec. A.2.1 for descriptions). They cover diverse surveillance environments, including indoor and outdoor scenes, public spaces, and crowded areas, with substantial variation in resolution, camera quality, and colour versus black-and-white imagery. 
SurFITR focuses on semantically grounded, localised manipulations across four editing types: {\it removal (RM)} (deleting an existing entity), {\it targeted replacement (RE(E))} (replacing an entity with a specific, context-consistent alternative), {\it open-ended replacement (RE(O))} (replacing an entity with a semantically related but not identical object), and {\it addition (ADD)} (inserting a new entity into the scene).
We explicitly apply tampering to both human and object targets, ensuring sufficient representation of each in surveillance settings. This results in over 100 distinct manipulation configurations. Full details of SurFITR are provided in Sec.~A.1, and extensive dataset visualisations are provided in Sec. D.

\subsection{Generation Pipeline}
\begin{figure}[t]
    \centering
    \includegraphics[width=0.95\linewidth]{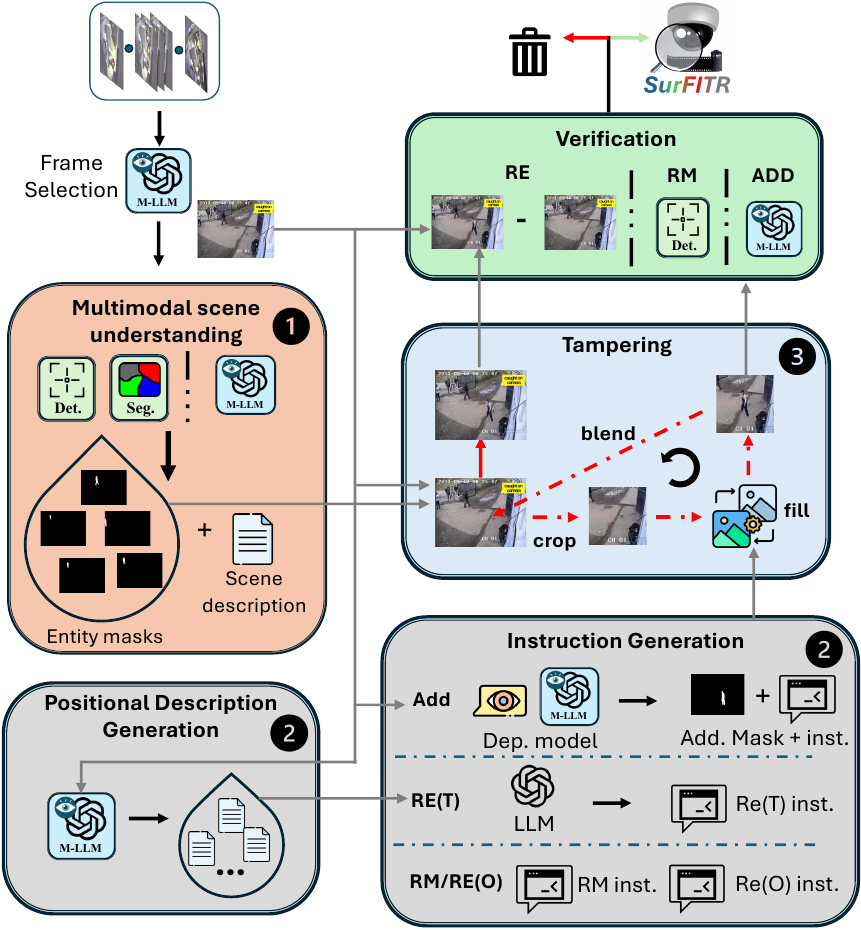}
    \vspace{-0.2cm}
    \caption{Overview of the SurFITR MLLM-powered tampering generation and verification pipeline. MLLMs are used to select forensically valuable frames from source corpora and, with assistance from grounding models, generate semantically and visually grounded tampering instructions. Localised, mask-guided manipulation is then applied, followed by verification to ensure tampering quality.}
    \label{fig:pipeline}
    \vspace{-0.6cm}
\end{figure}

To enable large-scale, fine-grained image tampering, we develop a fully automated multi-stage pipeline that generates semantically coherent edits while maintaining visual realism and contextual consistency. This design mimics real-world forgery, where specific high-value entities are selectively manipulated while the rest of the scene remains unchanged, making the manipulation less detectable. Due to potential misuse risks, we omit certain implementation details. We find that global, full-frame edits, even with grounding, remain comparatively easy to detect and are therefore unlikely to be used in realistic forgery scenarios.

As shown in Fig.~\ref{fig:pipeline}, The pipeline consists of three stages: (1) frame understanding and selection, which analyses scene content to identify suitable frames for manipulation; (2) tampering instruction generation, which produces context-aware and semantically grounded instructions for each frame; and (3) localised tampering, which performs spatially grounded, region-specific manipulation while preserving the remainder of the image. \smallskip

\noindent\textbf{Stage 1.} An open-vocabulary detector (YOLO-World \cite{cheng2024yolo}) and a promptable segmentation model (SAM 2.1 \cite{ravi2024sam}) are used to localise objects of interest (see Sec.~A.2.3 for details on open-world models). A MLLM (Qwen2.5-VL-72B \cite{qwen25vl}) generates structured scene descriptions capturing object attributes, spatial relationships, and lighting conditions. Each frame is assigned a forensic value score, and top-ranked frames are selected for manipulation. \smallskip

\noindent\textbf{Stage 2.} This stage generates textual instructions and visual guidance for image tampering. For {\it removal} and {\it open-ended replacement}, instructions are derived directly from the target mask and a category-level prompt. For targeted replacement and addition, a more structured reasoning process is required. 

For {\it targeted replacement}, scene descriptions are first summarised into structured format. An MLLM then generates detailed descriptions of the target object and its relationships with the surrounding context, capturing appearance attributes and spatial relationships. A text-based LLM reasons over these descriptions to propose context-consistent substitutions. This two-stage process produces more reliable suggestions than directly prompting an MLLM, likely due to the stronger reasoning capability of text-based LLMs.

For {\it addition}, an MLLM analyses each frame to jointly propose candidate objects and their placement locations, while enforcing physical plausibility (e.g., support surfaces and perspective consistency). Each candidate is scaled using a depth-aware process, where monocular depth maps (Depth Anything V2 \cite{depth_anything_v2}) are combined with category-level size statistics to ensure consistent scale relative to the scene. Realistic silhouette masks are generated by adapting same-category segmentation templates from SAM 2.1, replacing bounding boxes with natural contours, and serve as pixel-level masks for defining the tampering regions. We provide example instructions in Sec.~A.3. \smallskip

\noindent\textbf{Stage 3.} This stage performs localised, mask-controlled tampering and compositing. Tampering is restricted to pixel-level masks and their boundary regions, using patch-based manipulation (i.e., a local region enclosing the mask with some margin) while preserving the rest of the image. The patch is processed by an image generation model to produce a tampered version, which is then seamlessly composited back with blending constrained along the object silhouette, enabling fine-grained manipulation and precise tracking of tampered pixels for pixel-level ground truth generation. 
% To eliminate format-related shortcuts from diffusion and video encoding artifacts, all images are standardised to JPEG (quality 90).

\subsection{Quality Assurance}
\noindent\textbf{Verification.} To ensure dataset quality, we apply a task-specific verification process tailored to each tampering type. For removal, an open-vocabulary detector is reapplied within the target region to confirm the absence of the specified object or human. For open-ended and targeted replacement, a minimum pixel-level change threshold is enforced by comparing the tampered region with the original, filtering out trivial or failed edits. For addition, a MLLM model is used to verify the presence of the intended object category. These checks ensure that retained samples reflect the intended manipulation while minimising failed edits. The details of the verification is provided in Sec.~A.5.1.  \smallskip

\noindent\textbf{Quality Selection.} 
To further ensure dataset quality, we use Qwen2.5-VL-72B and Qwen3-VL-32B to rate each tampered image on two criteria: realism, for which higher scores indicate greater visual plausibility, and detectability, for which lower scores indicate fewer visible tampering artifacts. Both criteria are scored on a 10-point scale. Samples with realism below 5.5 or detectability above 4.5 are discarded. This step helps ensure that retained samples maintain reasonable visual realism and are not easily detectable. The details of quality selection is provided in Sec.~A.5.2. \smallskip

\noindent\textbf{Verified Subset.} To assess the gap between large-scale automated quality control and human verification, we construct a manually verified subset (over 5\% of the Base test set and 5\% of the Transfer set), retaining only samples with high visual quality and clearly identifiable tampering under ground-truth guidance. More discussions on the human verification is provided in Sec.~A.5.3.

\vspace{-0.2cm}
\section{Experiments}
\begin{table*}[t]
\centering
\resizebox{0.93\textwidth}{!}{%
\begin{tabular}{cc cc cc cc cc cc cc cc rr}
\toprule
\multirow{2}{*}{Task} & \multirow{2}{*}{Method}
  & \multicolumn{2}{c}{UCF}
  & \multicolumn{2}{c}{NTU}
  & \multicolumn{2}{c}{Shanghai}
  & \multicolumn{2}{c}{CUHK}
  & \multicolumn{2}{c}{UCSD\,1}
  & \multicolumn{2}{c}{UCSD\,2}
  & \multicolumn{2}{c}{Avg.}
  & \multicolumn{2}{c}{$\Delta$ vs.\ Verified} \\
\cmidrule(lr){3-4}\cmidrule(lr){5-6}\cmidrule(lr){7-8}
\cmidrule(lr){9-10}\cmidrule(lr){11-12}\cmidrule(lr){13-14}
\cmidrule(lr){15-16}\cmidrule(lr){17-18}
 & & AUROC & F1 & AUROC & F1 & AUROC & F1
   & AUROC & F1 & AUROC & F1 & AUROC & F1
   & AUROC & F1 & AUROC & F1 \\
\midrule
%-------- Detection --------
\multirow{10}{*}{{\textit{Det.}}}
  & PSCC-Net    & 0.315 & 0.300 & 0.424 & 0.463 & 0.577 & 0.536 & 0.666 & 0.668 & 0.937 & 0.007 & 0.800 & 0.030 & 0.620 & 0.334 & \textcolor{gray}{$+$0.011} & \textcolor{gray}{$+$0.003} \\
  & MVSSNet     & 0.495 & 0.667 & 0.470 & 0.667 & 0.448 & 0.667 & 0.436 & 0.667 & 0.546 & 0.667 & 0.526 & 0.667 & 0.487 & 0.667 & \textcolor{gray}{$-$0.035} & \textcolor{gray}{0.000} \\
  & TruFor      & 0.491 & 0.575 & 0.559 & 0.600 & 0.430 & 0.666 & 0.420 & 0.667 & 0.461 & 0.667 & 0.625 & 0.396 & 0.498 & 0.595 & \textcolor{gray}{$-$0.018} & \textcolor{gray}{$-$0.003} \\
  & HiFi-Net    & 0.537 & 0.000 & 0.504 & 0.001 & 0.546 & 0.039 & 0.277 & 0.000 & 0.661 & 0.000 & 0.944 & 0.000 & 0.578 & 0.007 & \textcolor{gray}{$-$0.002} & \textcolor{gray}{$-$0.001} \\
  & DRCT-2M     & 0.701 & 0.049 & 0.524 & 0.405 & 0.790 & 0.301 & 0.527 & 0.296 & 0.796 & 0.000 & 0.630 & 0.000 & 0.662 & 0.175 & \textcolor{gray}{$-$0.030} & \textcolor{gray}{$+$0.010} \\
  & Qwen2.5-VL-72B & 0.499 & 0.013 & 0.500 & 0.093 & 0.512 & 0.018 & 0.500 & 0.000 & 0.581 & 0.003 & 0.523 & 0.019 & 0.519 & 0.024 & \textcolor{gray}{$-$0.000} & \textcolor{gray}{$-$0.007} \\
  & Qwen3-VL-32B   & 0.550 & 0.060 & 0.398 & 0.008 & 0.564 & 0.001 & 0.548 & 0.000 & 0.501 & 0.002 & 0.568 & 0.007 & 0.521 & 0.013 & \textcolor{gray}{$+$0.002} & \textcolor{gray}{$+$0.001} \\
  & Qwen3-VL-8B    & 0.497 & 0.000 & 0.488 & 0.007 & 0.500 & 0.002 & 0.500 & 0.001 & 0.500 & 0.000 & 0.500 & 0.000 & 0.497 & 0.002 & \textcolor{gray}{$+$0.002} & \textcolor{gray}{$+$0.002} \\
  & ds-vl2   & 0.494 & 0.057 & 0.475 & 0.187 & 0.500 & 0.002 & 0.500 & 0.000 & 0.539 & 0.145 & 0.501 & 0.003 & 0.502 & 0.066 & \textcolor{gray}{$+$0.003} & \textcolor{gray}{$+$0.015} \\
  & Gemini-3   &0.492	&0.093	&0.511	&0.117	&0.497	&0.293	&0.507	&0.086	&0.499	&0.492	&0.502	&0.342	&0.501	&0.237	& \textcolor{gray}{$-$0.006} & \textcolor{gray}{$+$0.042}  \\
\midrule
%-------- Localisation --------
 & & P-IoU & P-F$_1$ & P-IoU & P-F$_1$ & P-IoU & P-F$_1$ & P-IoU & P-F$_1$ & P-IoU & P-F$_1$ & P-IoU & P-F$_1$ & P-IoU & P-F$_1$ & P-IoU & P-F$_1$  \\ \midrule
\multirow{7}{*}{{\textit{Loc.}}}
  & PSCC-Net     & 0.019 & 0.033 & 0.020 & 0.033 & 0.019 & 0.031 & 0.008 & 0.014 & 0.003 & 0.004 & 0.004 & 0.007 & 0.012 & 0.020 & \textcolor{gray}{$+$0.000} & \textcolor{gray}{$+$0.000} \\
  & CAT-Net      & 0.023 & 0.037 & 0.027 & 0.043 & 0.024 & 0.038 & 0.007 & 0.012 & 0.009 & 0.016 & 0.002 & 0.004 & 0.015 & 0.025 & \textcolor{gray}{$-$0.001} & \textcolor{gray}{$-$0.002} \\
  & MVSSNet      & 0.014 & 0.023 & 0.013 & 0.022 & 0.008 & 0.013 & 0.006 & 0.011 & 0.006 & 0.011 & 0.002 & 0.003 & 0.008 & 0.014 & \textcolor{gray}{$-$0.001} & \textcolor{gray}{$-$0.002} \\
  & HiFi-Net     & 0.000 & 0.000 & 0.000 & 0.000 & 0.000 & 0.000 & 0.000 & 0.000 & 0.000 & 0.000 & 0.000 & 0.000 & 0.000 & 0.000 & \textcolor{gray}{0.000}   & \textcolor{gray}{0.000} \\
  & ObjectFormer & 0.021 & 0.037 & 0.028 & 0.046 & 0.005 & 0.010 & 0.007 & 0.013 & 0.007 & 0.013 & 0.003 & 0.006 & 0.012 & 0.021 & \textcolor{gray}{$+$0.002} & \textcolor{gray}{$+$0.002} \\
  & IML-ViT      & 0.025 & 0.038 & 0.020 & 0.029 & 0.029 & 0.042 & 0.007 & 0.011 & 0.022 & 0.037 & 0.018 & 0.029 & 0.020 & 0.031 & \textcolor{gray}{$+$0.000} & \textcolor{gray}{$+$0.000} \\
  & TruFor       & 0.025 & 0.037 & 0.038 & 0.054 & 0.039 & 0.050 & 0.003 & 0.005 & 0.019 & 0.027 & 0.009 & 0.014 & 0.022 & 0.031 & \textcolor{gray}{$-$0.003} & \textcolor{gray}{$-$0.003} \\
\bottomrule
\end{tabular}%
}
\caption{Zero-shot detection and localisation on SurFITR-Base; $\Delta$ denotes the difference from the manually verified set.}
\label{tab:zs}
\vspace{-0.7cm}
\end{table*}

\begin{table}[t]
\centering
\begin{adjustbox}{max width=0.88\columnwidth}
\begin{tabular}{cccc|cccc}
\toprule
Task & Method & AUROC & F1 & Task  & Method & P-IoU & P-F1 \\
\midrule
\multirow{10}{*}{{Det.}}
 & PSCC-Net    & 0.665 & 0.384 & \multirow{7}{*}{{Loc.}} & PSCC-Net     & 0.714 & 0.305 \\
 & MVSSNet     & 0.477 & 0.667 & & CAT-Net      & 0.539 & 0.619 \\
 & TruFor      & 0.528 & 0.620 & & MVSSNet      & 0.571 & 0.574 \\
 & HiFi-Net    & 0.596 & 0.010 & & HiFi-Net     & 0.572 & 0.152 \\
 & DRCT-2M     & 0.578 & 0.165 & & ObjectFormer & 0.451 & 0.189 \\
 & Qwen2.5-VL-72b & 0.507 & 0.027 & & IML-ViT      & 0.408 & 0.149 \\
 & Qwen3-VL-32b   & 0.504 & 0.011 & & TruFor       & 0.399 & 0.136 \\
 & Qwen3-VL-8b   &0.497	&0.002 & &        &  &  \\
 & ds-vl2     & 0.507 & 0.086 & &              &       &       \\
 & Gemini-3   &0.503 &0.599 & &              &       &       \\
\bottomrule
\end{tabular}
\end{adjustbox}
\caption{Average performance on SurFITR Transfer.}
\label{tab:tzs}
\vspace{-0.9cm}
\end{table}

\subsection{Experimental Settings}
\noindent \textbf{Evaluation Metrics.} We report image-level detection performance using Accuracy (Acc), Area Under the ROC Curve (AUROC), and F1 score. For localisation, we evaluate pixel-level performance using pixel-wise Intersection over Union (P-IoU) and pixel-wise F1 score (P-F1), measuring the overlap between predicted manipulation masks and ground truth annotations (see Sec. B.1 for details). \smallskip \vspace{-0.3cm}

\noindent \textbf{Baseline methods.} We evaluate two categories of baselines: specialised forensic detectors and pretrained MLLMs, totalling 13 models. The forensic baselines include CNN- and transformer-based models for image forgery detection and localisation. Among these, PSCC-Net \cite{psccnet}, MVSSNet \cite{mvssnet}, and HiFi-Net \cite{hifinet} support both image-level detection and pixel-level localisation, while CAT-Net \cite{catnet}, ObjectFormer \cite{objectformer}, IML-ViT \cite{imlvit}, and TruFor \cite{trufor} are evaluated for localisation only. DRCT-2M \cite{drct} is additionally included for image-level classification. MLLMs assess image authenticity via visual–text reasoning and include both open-weight and commercial systems: Qwen3-VL-32B\cite{qwen3vl}, Qwen2.5-VL-72B\cite{qwen25vl}, Qwen3-VL-8B\cite{qwen3vl}, deepseek-vl2 (ds-vl2) \cite{wu2024deepseekvl2}, and Gemini Flash 3 \cite{gemini3_2025}. \smallskip 

\noindent \textbf{Evaluation Overview.} We use SurFITR to benchmark pretrained detection and localisation methods, and then evaluate the effect of training on SurFITR in both in-domain and cross-domain settings. For SurFITR as a test benchmark, we report zero-shot performance of pretrained models on both the Base and Transfer collections (Sec.~\ref{sec:test}). For SurFITR as training data (Sec.~\ref{sec:train}), we consider two settings: training on the full SurFITR Base training set and training on the UCF subset only. For models trained on the full set, we evaluate in-domain performance (Base Train $\rightarrow$ Base Test) and cross-domain performance with generation shifts (Base Train $\rightarrow$ Transfer). For models trained on the UCF subset, we evaluate cross-domain performance under two types of shift: dataset shift (Base UCF Train $\rightarrow$ Base Test) and combined dataset and generation shift (Base UCF Train $\rightarrow$ Transfer).\smallskip

\noindent \textbf{Implementation Details.} For zero-shot evaluation, we use the IMDL-BenCo~\cite{ma2024imdlbenco} implementations and the official codebases with pretrained weights. For fine-tuning, we adopt the same training code and train for 10 epochs. For Qwen3-VL-8B, we apply LoRA-based instruction tuning, where binary image-level labels and segmentation mask bounding boxes are included in the instruction-tuning outputs (see Sec.~B.2  for details).

\definecolor{diffbg}{gray}{0.92}
\newcommand{\D}[1]{\cellcolor{diffbg}#1}

\begin{table*}[t]
\centering
\label{tab:paper_new}
\resizebox{0.89\textwidth}{!}{%
\begin{tabular}{llllcccccccccccccc}
\toprule
\multirow{2}{*}{Task} & \multirow{2}{*}{Train Set} & \multirow{2}{*}{Method} & & \multicolumn{2}{c}{UCF} & \multicolumn{2}{c}{NTU} & \multicolumn{2}{c}{Shanghai} & \multicolumn{2}{c}{CUHK} & \multicolumn{2}{c}{UCSD 1} & \multicolumn{2}{c}{UCSD 2} & \multicolumn{2}{c}{Avg} \\
\cmidrule(lr){5-6}\cmidrule(lr){7-8}\cmidrule(lr){9-10}\cmidrule(lr){11-12}\cmidrule(lr){13-14}\cmidrule(lr){15-16}\cmidrule(lr){17-18}
& & & & AUROC & F$_1$ & AUROC & F$_1$ & AUROC & F$_1$ & AUROC & F$_1$ & AUROC & F$_1$ & AUROC & F$_1$ & AUROC & F$_1$ \\
\midrule
Det. & Base (all) & PSCC-Net & \textit{ft} & 0.991 & 0.927 & 0.933 & 0.846 & 1.000 & 0.988 & 0.999 & 0.814 & 1.000 & 0.298 & 1.000 & 0.731 & 0.987 & 0.767 \\
& & & \D{$\Delta$} & \D{{\color{teal}0.676}} & \D{{\color{teal}0.627}} & \D{{\color{teal}0.509}} & \D{{\color{teal}0.382}} & \D{{\color{teal}0.422}} & \D{{\color{teal}0.452}} & \D{{\color{teal}0.333}} & \D{{\color{teal}0.146}} & \D{{\color{teal}0.062}} & \D{{\color{teal}0.292}} & \D{{\color{teal}0.199}} & \D{{\color{teal}0.701}} & \D{{\color{teal}0.367}} & \D{{\color{teal}0.433}} \\
 &  & TruFor & \textit{ft} & 0.823 & 0.577 & 0.432 & 0.417 & 0.786 & 0.199 & 0.852 & 0.809 & 0.990 & 0.560 & 0.928 & 0.562 & 0.802 & 0.521 \\
& & & \D{$\Delta$} & \D{{\color{teal}0.326}} & \D{{\color{teal}0.041}} & \D{{\color{red}-0.130}} & \D{{\color{red}-0.153}} & \D{{\color{teal}0.344}} & \D{{\color{red}-0.463}} & \D{{\color{teal}0.431}} & \D{{\color{teal}0.142}} & \D{{\color{teal}0.545}} & \D{{\color{red}-0.108}} & \D{{\color{teal}0.307}} & \D{{\color{teal}0.559}} & \D{{\color{teal}0.304}} & \D{{\color{teal}0.003}} \\
 &  & HiFi-Net & \textit{ft} & 0.998 & 0.986 & 0.961 & 0.877 & 1.000 & 0.985 & 0.996 & 0.991 & 1.000 & 1.000 & 1.000 & 1.000 & 0.993 & 0.973 \\
& & & \D{$\Delta$} & \D{{\color{teal}0.479}} & \D{{\color{teal}0.318}} & \D{{\color{teal}0.462}} & \D{{\color{teal}0.210}} & \D{{\color{teal}0.503}} & \D{{\color{teal}0.318}} & \D{{\color{teal}0.496}} & \D{{\color{teal}0.324}} & \D{{\color{teal}0.500}} & \D{{\color{teal}0.333}} & \D{{\color{teal}0.500}} & \D{{\color{teal}0.333}} & \D{{\color{teal}0.490}} & \D{{\color{teal}0.306}} \\
 &  & Qwen3-VL-8B & \textit{ft} & 0.969 & 0.970 & 0.900 & 0.888 & 0.904 & 0.880 & 0.980 & 0.982 & 0.918 & 0.908 & 0.963 & 0.970 & 0.939 & 0.933 \\
& & & \D{$\Delta$} & \D{{\color{teal}0.490}} & \D{{\color{teal}0.652}} & \D{{\color{teal}0.438}} & \D{{\color{teal}0.678}} & \D{{\color{teal}0.402}} & \D{{\color{teal}0.562}} & \D{{\color{teal}0.483}} & \D{{\color{teal}0.657}} & \D{{\color{teal}0.418}} & \D{{\color{teal}0.575}} & \D{{\color{teal}0.463}} & \D{{\color{teal}0.636}} & \D{{\color{teal}0.449}} & \D{{\color{teal}0.627}} \\
\midrule
 & Base (UCF) & PSCC-Net & \textit{ft} & 0.982 & 0.866 & 0.524 & 0.392 & 0.441 & 0.313 & 0.774 & 0.129 & 0.997 & 0.860 & 0.949 & 0.716 & 0.778 & 0.546 \\
& & & \D{$\Delta$} & \D{{\color{teal}0.667}} & \D{{\color{teal}0.566}} & \D{{\color{teal}0.099}} & \D{{\color{red}-0.072}} & \D{{\color{red}-0.137}} & \D{{\color{red}-0.223}} & \D{{\color{teal}0.108}} & \D{{\color{red}-0.540}} & \D{{\color{teal}0.059}} & \D{{\color{teal}0.854}} & \D{{\color{teal}0.147}} & \D{{\color{teal}0.686}} & \D{{\color{teal}0.157}} & \D{{\color{teal}0.212}} \\
 &  & TruFor & \textit{ft} & 0.860 & 0.442 & 0.489 & 0.112 & 0.636 & 0.001 & 0.613 & 0.132 & 0.956 & 0.857 & 0.413 & 0.424 & 0.661 & 0.328 \\
& & & \D{$\Delta$} & \D{{\color{teal}0.362}} & \D{{\color{red}-0.094}} & \D{{\color{red}-0.073}} & \D{{\color{red}-0.458}} & \D{{\color{teal}0.194}} & \D{{\color{red}-0.661}} & \D{{\color{teal}0.192}} & \D{{\color{red}-0.535}} & \D{{\color{teal}0.511}} & \D{{\color{teal}0.188}} & \D{{\color{red}-0.209}} & \D{{\color{teal}0.421}} & \D{{\color{teal}0.163}} & \D{{\color{red}-0.190}} \\
 &  & HiFi-Net & \textit{ft} & 0.996 & 0.982 & 0.787 & 0.753 & 0.855 & 0.758 & 0.485 & 0.667 & 0.905 & 0.685 & 0.961 & 0.688 & 0.832 & 0.756 \\
& & & \D{$\Delta$} & \D{{\color{teal}0.478}} & \D{{\color{teal}0.315}} & \D{{\color{teal}0.287}} & \D{{\color{teal}0.086}} & \D{{\color{teal}0.358}} & \D{{\color{teal}0.092}} & \D{{\color{red}-0.015}} & \D{0.000} & \D{{\color{teal}0.405}} & \D{{\color{teal}0.019}} & \D{{\color{teal}0.461}} & \D{{\color{teal}0.021}} & \D{{\color{teal}0.329}} & \D{{\color{teal}0.089}} \\
 &  & Qwen3-VL-8B & \textit{ft} & 0.979 & 0.979 & 0.557 & 0.199 & 0.714 & 0.600 & 0.811 & 0.806 & 0.947 & 0.944 & 0.680 & 0.757 & 0.781 & 0.714 \\
& & & \D{$\Delta$} & \D{{\color{teal}0.483}} & \D{{\color{teal}0.979}} & \D{{\color{teal}0.069}} & \D{{\color{teal}0.192}} & \D{{\color{teal}0.214}} & \D{{\color{teal}0.598}} & \D{{\color{teal}0.311}} & \D{{\color{teal}0.805}} & \D{{\color{teal}0.447}} & \D{{\color{teal}0.944}} & \D{{\color{teal}0.180}} & \D{{\color{teal}0.757}} & \D{{\color{teal}0.284}} & \D{{\color{teal}0.712}} \\
\midrule
& & & & P-IoU & P-F$_1$ & P-IoU & P-F$_1$ & P-IoU & P-F$_1$ & P-IoU & P-F$_1$ & P-IoU & P-F$_1$ & P-IoU & P-F$_1$ & P-IoU & P-F$_1$ \\
\midrule
Loc. & Base (all) & PSCC-Net & \textit{ft} & 0.068 & 0.102 & 0.070 & 0.104 & 0.062 & 0.093 & 0.036 & 0.057 & 0.012 & 0.017 & 0.013 & 0.018 & 0.043 & 0.065 \\
& & & \D{$\Delta$} & \D{{\color{teal}0.049}} & \D{{\color{teal}0.070}} & \D{{\color{teal}0.050}} & \D{{\color{teal}0.071}} & \D{{\color{teal}0.043}} & \D{{\color{teal}0.062}} & \D{{\color{teal}0.029}} & \D{{\color{teal}0.043}} & \D{{\color{teal}0.009}} & \D{{\color{teal}0.013}} & \D{{\color{teal}0.008}} & \D{{\color{teal}0.011}} & \D{{\color{teal}0.031}} & \D{{\color{teal}0.045}} \\
 &  & TruFor & \textit{ft} & 0.168 & 0.205 & 0.197 & 0.239 & 0.121 & 0.149 & 0.068 & 0.087 & 0.030 & 0.037 & 0.043 & 0.054 & 0.105 & 0.129 \\
& & & \D{$\Delta$} & \D{{\color{teal}0.142}} & \D{{\color{teal}0.166}} & \D{{\color{teal}0.156}} & \D{{\color{teal}0.182}} & \D{{\color{teal}0.083}} & \D{{\color{teal}0.098}} & \D{{\color{teal}0.064}} & \D{{\color{teal}0.082}} & \D{{\color{teal}0.011}} & \D{{\color{teal}0.010}} & \D{{\color{teal}0.035}} & \D{{\color{teal}0.042}} & \D{{\color{teal}0.082}} & \D{{\color{teal}0.097}} \\
 &  & HiFi-Net & \textit{ft} & 0.020 & 0.036 & 0.024 & 0.039 & 0.002 & 0.004 & 0.009 & 0.018 & 0.002 & 0.004 & 0.001 & 0.002 & 0.010 & 0.017 \\
& & & \D{$\Delta$} & \D{{\color{teal}0.020}} & \D{{\color{teal}0.036}} & \D{{\color{teal}0.024}} & \D{{\color{teal}0.039}} & \D{{\color{teal}0.002}} & \D{{\color{teal}0.004}} & \D{{\color{teal}0.009}} & \D{{\color{teal}0.018}} & \D{{\color{teal}0.002}} & \D{{\color{teal}0.004}} & \D{{\color{teal}0.001}} & \D{{\color{teal}0.002}} & \D{{\color{teal}0.010}} & \D{{\color{teal}0.017}} \\
 &  & Qwen3-VL-8B & \textit{ft} & 0.116 & 0.172 & 0.075 & 0.116 & 0.055 & 0.088 & 0.100 & 0.151 & 0.053 & 0.079 & 0.086 & 0.134 & 0.081 & 0.123 \\
& & & \D{$\Delta$} & \D{{\color{teal}0.116}} & \D{{\color{teal}0.172}} & \D{{\color{teal}0.068}} & \D{{\color{teal}0.105}} & \D{{\color{teal}0.055}} & \D{{\color{teal}0.088}} & \D{{\color{teal}0.100}} & \D{{\color{teal}0.151}} & \D{{\color{teal}0.053}} & \D{{\color{teal}0.079}} & \D{{\color{teal}0.086}} & \D{{\color{teal}0.134}} & \D{{\color{teal}0.080}} & \D{{\color{teal}0.121}} \\
\midrule
 & Base (UCF) & PSCC-Net & \textit{ft} & 0.038 & 0.060 & 0.008 & 0.012 & 0.003 & 0.004 & 0.001 & 0.002 & 0.010 & 0.017 & 0.005 & 0.008 & 0.011 & 0.017 \\
& & & \D{$\Delta$} & \D{{\color{teal}0.019}} & \D{{\color{teal}0.027}} & \D{{\color{red}-0.013}} & \D{{\color{red}-0.021}} & \D{{\color{red}-0.016}} & \D{{\color{red}-0.027}} & \D{{\color{red}-0.006}} & \D{{\color{red}-0.012}} & \D{{\color{teal}0.007}} & \D{{\color{teal}0.013}} & \D{{\color{teal}0.001}} & \D{{\color{teal}0.001}} & \D{{\color{red}-0.001}} & \D{{\color{red}-0.003}} \\
 &  & TruFor & \textit{ft} & 0.165 & 0.200 & 0.055 & 0.069 & 0.018 & 0.023 & 0.004 & 0.005 & 0.040 & 0.049 & 0.036 & 0.048 & 0.053 & 0.066 \\
& & & \D{$\Delta$} & \D{{\color{teal}0.138}} & \D{{\color{teal}0.162}} & \D{{\color{teal}0.014}} & \D{{\color{teal}0.013}} & \D{{\color{red}-0.021}} & \D{{\color{red}-0.028}} & \D{{\color{teal}0.001}} & \D{{\color{teal}0.000}} & \D{{\color{teal}0.021}} & \D{{\color{teal}0.022}} & \D{{\color{teal}0.028}} & \D{{\color{teal}0.036}} & \D{{\color{teal}0.030}} & \D{{\color{teal}0.034}} \\
 &  & HiFi-Net & \textit{ft} & 0.020 & 0.036 & 0.015 & 0.025 & 0.001 & 0.002 & 0.008 & 0.015 & 0.005 & 0.009 & 0.003 & 0.006 & 0.009 & 0.016 \\
& & & \D{$\Delta$} & \D{{\color{teal}0.020}} & \D{{\color{teal}0.036}} & \D{{\color{teal}0.015}} & \D{{\color{teal}0.025}} & \D{{\color{teal}0.001}} & \D{{\color{teal}0.002}} & \D{{\color{teal}0.008}} & \D{{\color{teal}0.015}} & \D{{\color{teal}0.005}} & \D{{\color{teal}0.009}} & \D{{\color{teal}0.003}} & \D{{\color{teal}0.006}} & \D{{\color{teal}0.009}} & \D{{\color{teal}0.016}} \\
 &  & Qwen3-VL-8B & \textit{ft} & 0.116 & 0.170 & 0.003 & 0.006 & 0.001 & 0.001 & 0.008 & 0.014 & 0.001 & 0.002 & 0.025 & 0.039 & 0.026 & 0.039 \\
& & & \D{$\Delta$} & \D{{\color{teal}0.116}} & \D{{\color{teal}0.170}} & \D{{\color{red}-0.004}} & \D{{\color{red}-0.005}} & \D{{\color{teal}0.001}} & \D{{\color{teal}0.001}} & \D{{\color{teal}0.008}} & \D{{\color{teal}0.014}} & \D{{\color{teal}0.001}} & \D{{\color{teal}0.002}} & \D{{\color{teal}0.025}} & \D{{\color{teal}0.039}} & \D{{\color{teal}0.025}} & \D{{\color{teal}0.037}} \\
\bottomrule
\end{tabular}%
}
\caption{Finetuned results on SurFITR Base. $\Delta$: gain over zero-shot; green: improvement; red: degradation.}
\label{tab:ft}
\vspace{-0.7cm}
\end{table*}

\subsection{SurFITR as a Test Benchmark}
\label{sec:test}
\subsubsection{Detection on SurFITR Base.} 
Table \ref{tab:zs} (Det.) reports image-level detection performance across the six source subsets. Overall, existing methods struggle on SurFITR, particularly in terms of F1 score. While several methods achieve moderate AUROC values (often above 0.5 and up to 0.7), their F1 scores remain extremely low, indicating poor score calibration likely caused by domain gaps and ineffective threshold-based detection. This suggests that models trained on existing datasets learn domain-specific features that do not transfer to surveillance imagery, resulting in unreliable instance-level detection. Performance also varies substantially across subsets, further indicating limited generalisation across surveillance environments and manipulation contexts.

\subsubsection{Localisation on SurFITR Base.} Table \ref{tab:zs} (Loc.) reports pixel-level localisation performance across the six subsets using P-IoU and P-F1. Overall, all methods exhibit very limited localisation capability, with extremely low scores across both metrics. Even the best-performing method, TruFor, achieves fairly low performance, and most approaches produce near-zero results on multiple datasets, indicating that existing localisation models struggle to capture subtle, localised tampering in surveillance imagery, consistent with the detection results and the domain gap discussed above.

\subsubsection{Performance on SurFITR Transfer.} We report average detection and localisation performance in Tab.\ref{tab:tzs}, with full results in Tab.~8 (Sec.~C.2.2). Trends are similar to the Base subset, but with non-trivial performance gaps. Despite differences in data composition, results show that both scene semantics and generation methods affect performance, highlighting the importance of diverse SOTA models for cross-model generalisation.

\subsubsection{Performance on the Verified Set} 
We report the performance difference between the verified subset and the full test set in Tab.~7 (Sec.~C2.1) to assess consistency between large-scale evaluation and a manually verified subset with stricter quality control. The differences are small, generally below 5\% and mostly within 1\%, indicating that large-scale automated verification ensures samples with similar detectability to the human-verified subset.

\subsection{SurFITR as Training Data}
\label{sec:train}
We train selected baseline methods on the full SurFITR Base training set (Sec.~\ref{sec:ft_all}) and on the UCF subset of the Base training set (Sec.~\ref{sec:ft_ucf}), for their in-domain and cross-domain performance.

\subsubsection{Fine-tuned using SurFITR Base (all)}
\label{sec:ft_all}
We evaluate performance under an in-domain setting (Base Train $\rightarrow$ Base Test) and a cross-domain setting involving different image generation models (Base Train $\rightarrow$ Transfer). \smallskip

\noindent \textbf{In-domain performance.} Tab.~\ref{tab:ft} (Base, all) reports the performance of selected baselines on the Base test set. Fine-tuning improves detection across datasets, with PSCC-Net, HiFi-Net, and Qwen3-VL-8B achieving very high AUROC and F1, while TruFor shows more variability but still improves overall. Notably, the SurFITR-tuned Qwen3-VL-8B achieves performance comparable to dedicated detectors, suggesting that SurFITR can serve as an effective training resource for improving forensic capability in MLLMs. For localisation, all models improve after fine-tuning. TruFor shows the largest gains, while HiFi-Net exhibits only minor improvements from a near-zero pretrained baseline. These results indicate that SurFITR provides informative supervision for adapting models to surveillance-style manipulations. Notably, models display different strengths in detection and localisation, suggesting that both tasks should be considered jointly when evaluating model capability. \smallskip

\begin{table}[t]
\centering
\resizebox{\columnwidth}{!}{%
\begin{tabular}{llcccccccc}
\toprule
& & \multicolumn{4}{c}{Detection} & \multicolumn{4}{c}{Localisation} \\
\cmidrule(lr){3-6}\cmidrule(lr){7-10}
& & \multicolumn{2}{c}{Base (all)} & \multicolumn{2}{c}{Base (UCF)} & \multicolumn{2}{c}{Base (all)} & \multicolumn{2}{c}{Base (UCF)} \\
\cmidrule(lr){3-4}\cmidrule(lr){5-6}\cmidrule(lr){7-8}\cmidrule(lr){9-10}
Method & & AUROC & F$_1$ & AUROC & F$_1$ & P-IoU & P-F$_1$ & P-IoU & P-F$_1$ \\
\midrule
PSCC-Net & \textit{ft} & 0.952 & 0.575 & 0.793 & 0.548 & 0.058 & 0.083 & 0.018 & 0.028 \\
& \D{$\Delta$} & \D{{\color{teal}0.286}} & \D{{\color{teal}0.191}} & \D{{\color{teal}0.128}} & \D{{\color{teal}0.164}} & \D{{\color{teal}0.031}} & \D{{\color{teal}0.039}} & \D{{\color{red}-0.010}} & \D{{\color{red}-0.015}} \\
TruFor & \textit{ft} & 0.785 & 0.502 & 0.655 & 0.351 & 0.115 & 0.139 & 0.081 & 0.100 \\
& \D{$\Delta$} & \D{{\color{teal}0.257}} & \D{{\color{red}-0.118}} & \D{{\color{teal}0.127}} & \D{{\color{red}-0.269}} & \D{{\color{teal}0.047}} & \D{{\color{teal}0.052}} & \D{{\color{teal}0.013}} & \D{{\color{teal}0.014}} \\
HiFi-Net & \textit{ft} & 0.981 & 0.953 & 0.820 & 0.744 & 0.012 & 0.021 & 0.012 & 0.020 \\
& \D{$\Delta$} & \D{{\color{teal}0.385}} & \D{{\color{teal}0.942}} & \D{{\color{teal}0.224}} & \D{{\color{teal}0.734}} & \D{{\color{red}-0.035}} & \D{{\color{red}-0.031}} & \D{{\color{teal}0.012}} & \D{{\color{teal}0.020}} \\
Qwen3-VL-8B  & \textit{ft} & 0.939 & 0.933 & 0.793 & 0.735 & 0.124 & 0.184 & 0.039 & 0.058 \\
& \D{$\Delta$} & \D{{\color{teal}0.442}} & \D{{\color{teal}0.931}} & \D{{\color{teal}0.295}} & \D{{\color{teal}0.732}} & \D{{\color{teal}0.122}} & \D{{\color{teal}0.182}} & \D{{\color{teal}0.037}} & \D{{\color{teal}0.055}} \\
\bottomrule
\end{tabular}%
\vspace{-0.4cm}
}
\caption{Avg.    Finetuned performance on SurFITR-Transfer.}
\label{tab:ft_trans}
\vspace{-0.7cm}
\end{table}
\noindent \textbf{Cross-domain performance.} Tab.~\ref{tab:ft_trans} (Base, all) shows the performance of the same baselines on SurFITR-Transfer, where similar trends are observed. Although direct comparison is not possible due to differences in data composition, the overall performance remains comparable, with only slight degradation. This indicates that switching to different or more recent generation models has limited impact, and that the learned cues generalise across generation methods under the same editing settings.

\subsubsection{Fine-tuned using SurFITR Base (UCF only)} 
\label{sec:ft_ucf}

We evaluate UCF-trained baselines under two cross-domain settings: Base (UCF) → Base (cross-dataset) and Base (UCF) → Transfer (cross-dataset and generation model). Average results are reported in Table~\ref{tab:ft_trans}, with full results in Sec.~C.3. \smallskip

\noindent \textbf{Cross-dataset performance} Tab.~\ref{tab:ft} (Base, UCF) reports fine-tuned performance and improvements over zero-shot when training on UCF samples from the SurFITR Base partition. Compared to full fine-tuning, gains are lower and vary across datasets, indicating scene-specific domain gaps and the importance of scene diversity. For detection, improvements remain significant overall, though TruFor shows a notable drop in F1 on some datasets. For localisation, TruFor achieves non-trivial gains, while other methods remain close to zero-shot on unseen datasets. These results suggest that SurFITR provides useful supervision for cross-scene generalisation, while baseline models differ in their ability to exploit it across tasks. \smallskip

\noindent \textbf{Cross-domain and Cross-model Performance.} 
Tab.\ref{tab:ft_trans} (Base (UCF) columns) shows cross-domain performance across both datasets and generation models. The trend is similar to cross-dataset evaluation, but with moderately lower gains. This suggests that unseen generation models introduce additional complexity, and that current baseline methods still learn model-specific features and remain sensitive to such cues under surveillance settings. Further analysis of performance across editing models and types is provided in Sec.~C1. \smallskip

\noindent \textbf{Open Research Questions.} Even after training on SurFITR, detection performance remains sensitive to scene variation (Fig.~5), with noticeable degradation under cross-dataset and cross-domain settings, while localisation performance is strongly influenced by manipulation size (Fig.~6), with smaller edits being substantially more challenging (see Sec.~C.1 for details). These observations reveal systematic limitations of current methods on surveillance-style data, where models fail to maintain consistent detection across scenes and struggle to localise fine-grained manipulations, demonstrating the value of SurFITR as a testbed for studying these challenges and enabling the development of more robust models.

\section{Conclusion}
We introduce SurFITR, a novel dataset for surveillance-style image forgery detection and localisation capturing fine-grained, localised tampering across diverse real-world scenes. Experiments show that existing methods degrade significantly under subtle surveillance manipulations, while training on SurFITR yields substantial improvements. Notable gaps remain in cross-domain generalisation and the localisation of subtle manipulations, positioning SurFITR as both a benchmark and a foundation for developing specialised forensic models for surveillance imagery. \medskip

\noindent\textbf{Ethical Considerations and Limitations.} SurFITR is designed to support forensic research rather than enable forgery. To mitigate potential misuse, we omit key implementation details of the generation pipeline and release the dataset under a research-only license. All source images are drawn from publicly available datasets, and the scene distribution may not fully capture private or sensitive environments commonly encountered in real-world reporting, which are inaccessible due to privacy constraints. The underlying image generation models are already publicly available, and SurFITR aims to facilitate the development of methods for detecting and mitigating their misuse.

% \begin{acks}
% \end{acks}


\bibliographystyle{ACM-Reference-Format}
\bibliography{sample-base}

\label{apd:vis}

\end{document}